\documentclass[letterpaper, 10 pt, journal, twoside]{ieeetran}
\usepackage{amsmath,amsfonts}
\usepackage{algorithm}
\usepackage{array}
\usepackage[caption=false,font=normalsize,labelfont=sf,textfont=sf]{subfig}
\usepackage{textcomp}
\usepackage{stfloats}
\usepackage{url}
\usepackage{verbatim}
\usepackage{graphicx}
\usepackage{cite}
\usepackage{graphicx}
\usepackage[table]{xcolor}
\usepackage{multirow}
\usepackage{amsmath}
\definecolor{MyLightGreen}{HTML}{50F22C}
\usepackage{booktabs}
\usepackage{tabularx}
\usepackage{makecell}
\usepackage{xcolor} 
\usepackage{subcaption}
\usepackage{cite}
\usepackage{algpseudocode} 
\usepackage{amssymb}
\usepackage{adjustbox}
\definecolor{light_blue}{rgb}{0.9, 0.95, 1.0}
\definecolor{med_blue}{rgb}{0.75, 0.9, 1.0}
\definecolor{dark_blue}{rgb}{0.5, 0.75, 1.0}
\definecolor{deep_blue}{rgb}{0.3, 0.6, 1.0}
\definecolor{very_deep_blue}{rgb}{0.1, 0.4, 1.0}

\begin{document}

\title{X-IONet: Cross-Platform Inertial Odometry Network \\ for Pedestrian and Legged Robot}

\markboth{IEEE Robotics and Automation Letters. Preprint Version. Accepted April, 2026}
{Shen \MakeLowercase{\textit{et al.}}: X-IONet: Cross-Platform Inertial Odometry Network for Pedestrian and Legged Robot} 

\author{Dehan Shen and Changhao Chen%
\thanks{
This work was supported by National Natural Science Foundation of China under Grant 62573370 and Education Department of Guangdong Province under grant 2025ZDZX3051. (Corresponding author: Changhao Chen)} 
\thanks{Dehan Shen is with Intelligent Transportation Thrust, The Hong Kong University of Science and Technology
(Guangzhou), Guangzhou, 511453, China {\tt\footnotesize (email: dehanshen@hkust-gz.edu.cn)}}%
\thanks{Changhao Chen is with Intelligent Transportation Thrust and Artificial Intelligence Thrust, The Hong Kong University of Science and Technology (Guangzhou), Guangzhou 511453, China, and also with  the Division of Emerging Interdisciplinary Areas, The Hong Kong University of Science and Technology, Hong Kong SAR 999077, China {\tt\footnotesize (email: changhaochen@hkust-gz.edu.cn)}}%

}




\maketitle

\begin{abstract}
Learning-based inertial odometry has achieved remarkable progress in pedestrian navigation. However, extending these methods to quadruped robots remains challenging due to their distinct and highly dynamic motion patterns. Models that perform well on pedestrian data often experience severe degradation when deployed on legged platforms.
To tackle this challenge, we introduce X-IONet, a cross-platform inertial odometry framework that operates solely using a single Inertial Measurement Unit (IMU).
X-IONet incorporates a rule-based expert selection module to classify motion platforms and route IMU sequences to platform-specific expert networks. The displacement prediction network features a dual-stage attention architecture that jointly models long-range temporal dependencies and inter-axis correlations, enabling accurate motion representation. It outputs both displacement and associated uncertainty, which are further fused through an Extended Kalman Filter (EKF) for robust state estimation.
Extensive experiments on the public RoNIN pedestrian dataset, the GrandTour quadruped dataset, and a self-collected Go2 quadruped dataset demonstrate that X-IONet achieves state-of-the-art performance, reducing ATE and RTE by 14.3\% and 11.4\% on RoNIN, 11.8\% and 9.7\% on GrandTour, and 52.8\% and 41.3\% on Go2.
These results highlight X-IONet’s effectiveness for accurate and robust inertial navigation across both human and legged robot platforms.
\end{abstract}

\begin{IEEEkeywords}
Inertial Navigation, Localization, Pedestrian Navigation, Deep Neural Network
\end{IEEEkeywords}

\section{Introduction}
\IEEEPARstart{I}{nertial} navigation estimates ego-motion and position solely from Inertial Measurement Unit (IMU) data. Unlike visual–inertial odometry \cite{VIO, vio2025, buchanan2022deep}, lidar–inertial odometry \cite{LIO, lidar2024, LIO2024}, or GNSS-based localization \cite{GNSS, GNSS2024}, inertial navigation is fully self-contained and does not rely on external signals \cite{Survey}. This independence enables robust performance in visually degraded environments and GNSS-denied settings. However, unavoidable sensor noise causes significant drift during long-term pure-inertial integration, making accurate standalone inertial navigation a long-standing challenge.

Recently, deep learning has shown great potential to enhance inertial odometry. Learning-based methods leverage neural architectures to directly extract motion patterns from IMU signals, effectively constraining dead-reckoning drift and improving state estimation accuracy. These approaches have been applied to pedestrian motion \cite{IONet, TLIO, IMUNet, CTIN, iMoT, RoNIN, deepils}, legged robots \cite{learningRobot}, and unmanned aerial vehicles (UAVs) \cite{AirIO, IMO, DIDO, UAV_dead-reckoning}.

\begin{figure}
    \centering
    \captionsetup{font=small}
    \includegraphics[width=\linewidth]{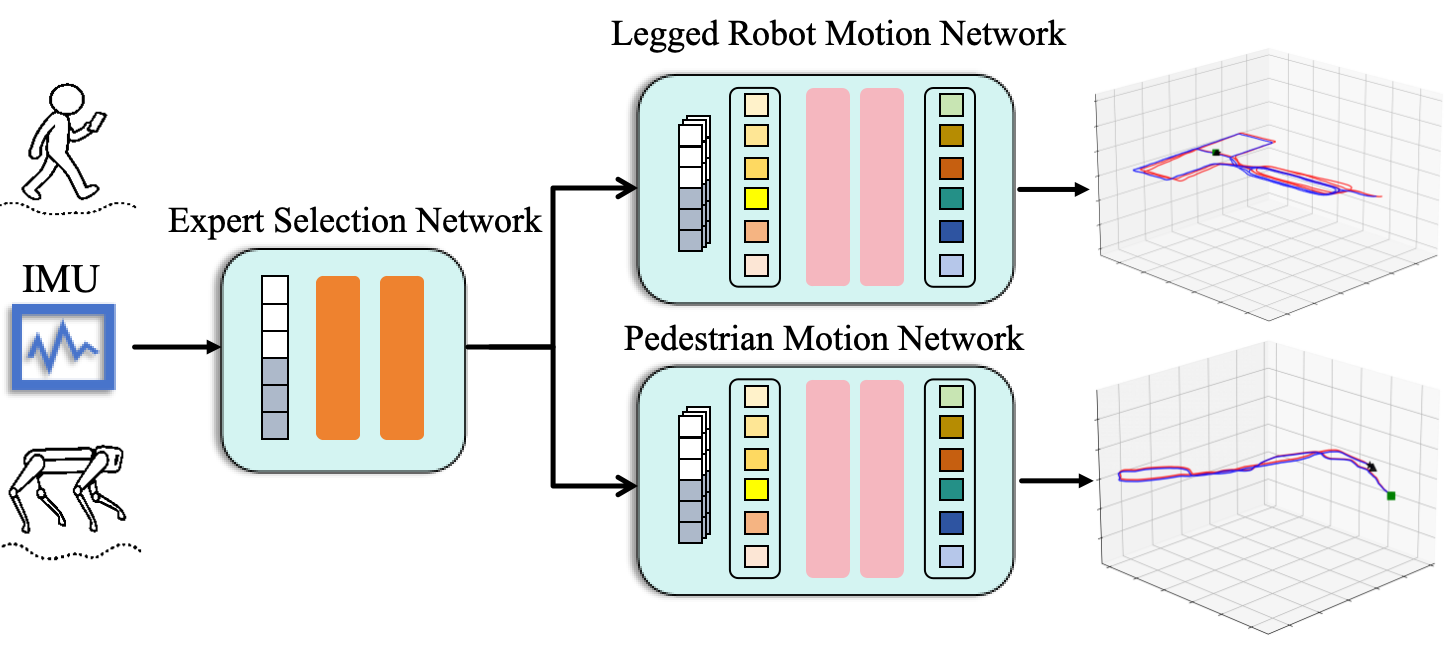}
    \caption{Overview of the proposed X-IONet model. IMU sequences are first processed by an expert selection network, which identifies the motion platform and routes the input to the corresponding platform-specific motion network. Each motion expert network subsequently predicts the motion trajectory for its designated platform.}
    \label{concept}
\end{figure}

Despite promising progress, most existing methods are tailored to a single platform, leaving cross-platform inertial odometry largely unsolved. Although pedestrian and quadruped locomotion share certain characteristics—such as periodicity and gravity-related priors—quadruped motion is considerably more dynamic and diverse. As shown in Fig.~\ref{concept}, quadrupeds perform rapid accelerations and decelerations, lateral and backward movements, and highly non-periodic maneuvers, producing complex inertial signatures absent in human motion. Moreover, inertial motion cues are inherently multi-scale and multi-dimensional; existing architectures struggle to extract and fuse such features effectively, resulting in inaccurate displacement predictions. These challenges call for new neural models that capture both long-range temporal dependencies and cross-axis correlations in IMU data.

To address these challenges, we propose \textbf{X-IONet} (\textbf{Cross}-Platform \textbf{I}nertial \textbf{O}dometry \textbf{Net}work), a novel and generalizable deep learning framework for inertial odometry estimation across both human and quadruped robotic platform. As shown in Fig. \ref{concept}, X-IONet incorporates a rule-based expert selection module that routes IMU data to specialized expert networks based on platform characteristics. Each expert network adopts a dual-stage attention–based displacement prediction model, capturing long-range temporal dependencies via temporal self-attention and inter-axis relationships via dimensional self-attention. The network predicts motion displacement and corresponding uncertainty, and integrates these estimates into an Extended Kalman Filter (EKF) for robust motion state estimation.
Extensive experiments on the RoNIN\cite{RoNIN} pedestrian dataset, the GrandTour\cite{grandtour} quadruped dataset, and our self-collected Go2 quadruped dataset demonstrate that X-IONet achieves state-of-the-art accuracy, generalizes effectively across platforms, and maintains robustness under distinct motion patterns.

The main contributions of this paper are as follows:
\begin{itemize}

    \item We propose X-IONet, a generalizable cross-platform inertial odometry network that estimates displacement and uncertainty from raw IMU data, enabling robust localization for pedestrian and quadruped robots.
    \item We introduce a novel displacement prediction network with dual-stage attention mechanism that hierarchically captures multi-scale temporal dependencies and inter-axis correlations in inertial signals, providing a more expressive and physically consistent representation for inertial odometry estimation.
    \item We perform comprehensive evaluations on public pedestrian and quadruped datasets, as well as a self-collected quadruped dataset, showing that X-IONet generalizes across distinct motion patterns and achieves state-of-the-art inertial navigation performance.

\end{itemize}

\section{RELATED WORK}
Recent advances in deep learning–based inertial odometry focus on modeling raw IMU signals for accurate motion estimation. IONet \cite{IONet} first demonstrated the feasibility of learning motion patterns directly from raw inertial data using recurrent neural networks. RoNIN \cite{RoNIN} further advanced this direction by proposing three end-to-end frameworks based on Residual Networks (ResNet) \cite{ResNet}, Long Short-Term Memory (LSTM) networks \cite{LSTM}, and Temporal Convolutional Networks (TCN) \cite{TCN}, and also released a large-scale dataset for pedestrian inertial navigation. With the success of Transformer architectures in natural language processing \cite{BERT} and computer vision \cite{ViT}, CTIN \cite{CTIN} introduced a Transformer-based model for inertial data, enabling global motion pattern modeling through dynamic attention computation. Building upon this idea, iMoT \cite{iMoT} incorporated local self-attention and convolutional layers, improving the extraction of fine-grained local motion features. Beyond attention mechanisms, RIO \cite{RIO} introduced rotationally equivariant networks for self-supervised pose estimation, while EqNIO \cite{EqNIO} learned rotation-invariant displacement priors through canonical frame mapping. Jayanth et al.\cite{LieEvent} proposed using Lie Events as an alternative to conventional raw IMU data as inputs to the Neural Differential Pose (NDP) model. Li et al.\cite{Liyan} modeled the dominance of different signals under various motion patterns and generated virtual training samples based on this modeling, thereby alleviating the problem of insufficient labeled data. These efforts highlight rapid progress in pedestrian inertial odometry, but deep learning for quadruped robots using only IMU remains underexplored. 

The Extended Kalman Filter (EKF) is widely used for nonlinear state estimation.
Early works applied EKF to quadruped robot state estimation by fusing IMU and kinematic data. For example, Lin et al.\cite{RobotEKF} fused IMU and hexapod kinematics but assumed that three legs remained in contact with the ground, limiting generality. Later, Bloesch et al.\cite{MultisensorRobot} combined IMU, joint encoders, and foot-contact sensors within an EKF, while Yang et al.\cite{multiIMURobot} proposed mounting additional IMUs on the legs and fusing them with body-mounted IMU and encoder data, enabling more accurate estimation of both body and foot states. While effective, these methods depend on multiple heterogeneous sensors, which increase hardware and computational complexity. 
TLIO \cite{TLIO} bridged learning and filtering by feeding neural network–predicted displacement and uncertainty into an EKF, significantly improving the robustness of IMU-only odometry. This paradigm has since been extended to pedestrian and UAV navigation \cite{AirIO, DIDO, IMO, UAV_dead-reckoning}, demonstrating the benefits of combining data-driven models with probabilistic filtering. In the quadruped domain, Buchanan et al.\cite{learningRobot} integrated neural network–based state predictions with joint encoder measurements within an EKF for improved estimation. However, the approach relied heavily on encoder data and introduced little innovation in network design, which reduces generality and practical deployment.

\begin{figure*}[!t]
    \centering
    \captionsetup{font=small}
    \includegraphics[width=\linewidth]{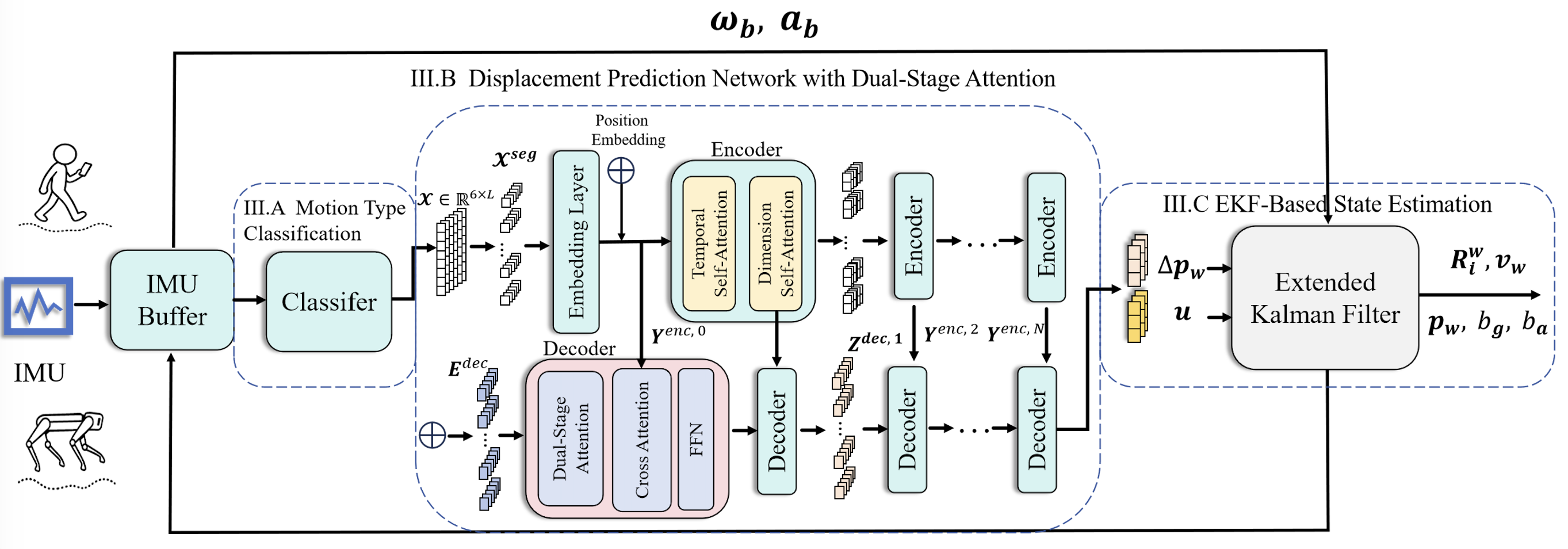}
    \caption{The framework of the proposed X-IONet model. The raw inertial data are rotated using the attitude estimated by EKF, so that the input is represented in a gravity-aligned local frame. The transformed IMU sequence is then fed into a rule-based expert selection network to identify the motion platform. The data are routed to a displacement prediction network to regress displacement and uncertainty. The predicted results are refined through the EKF to achieve accurate and robust cross-platform inertial odometry.}
    \label{fig:overall framework}
\end{figure*}

\section{METHODOLOGY}
This section presents the proposed X-IONet framework, which consists of three key components: (i) a rule-based expert selection network, (ii) a displacement prediction network, and (iii) an Extended Kalman Filter (EKF)–based state estimator. The overall pipeline is shown in Fig.~\ref{fig:overall framework}. First, the expert selection network identifies the motion platform via a lightweight 1-D convolutional classifier and routes the IMU sequence to the appropriate pre-trained expert model (Section~\ref{sec:classifier}). Then, the displacement prediction network estimates motion displacement and uncertainty using a dual-stage attention architecture that jointly captures temporal and inter-axis dependencies (Section~\ref{sec:displacement}). Finally, an EKF integrates the predicted displacement and uncertainty to refine global state estimation (Section~\ref{sec:EKF}).

\subsection{Rule-based Expert Selection Network}
\label{sec:classifier}
Human and quadruped motion generate distinct inertial signatures. To enable cross-platform deployment, we introduce a rule-based expert selection network consisting of a 1-D CNN classifier and a platform-aware routing module. Given an IMU sequence, the classifier predicts the platform type, and the routing module directs the sequence to the corresponding expert model trained for that platform.

\textbf{Classification network.} 
The classification network adopts a 1-D convolutional structure to identify the platform type.
The input $\mathcal{X} \in \mathbb{R} ^{6 \times T \times f }$ represents a segment of IMU data, where $T$ denotes the time window and f denotes the sampling frequency. Each time step includes six features—three from the gyroscope ($\boldsymbol{\omega}\in \mathbb{R}^3$) and three from the accelerometer ($\mathbf{a}\in \mathbb{R}^3$).

Feature extraction is performed using a three-layer stacked 1-D convolutional block, where each layer is formulated as:
\begin{equation}
\boldsymbol{\mathcal{X}}_{n+1}=MaxPool(relu(BN(Conv1d(\boldsymbol{\mathcal{X}}_n))))
\end{equation}
where $\boldsymbol{\mathcal{X}}_n$ is the input for each layer, and it is the IMU data segment in the first layer. $Conv1d$ is the 1-D convolution operation; $relu( \cdot  )$ is an activation function; $BN( \cdot  ) $ is batch normalization; $MaxPool( \cdot  )$ is a max pooling layer. 

\begin{figure}[h]
    \centering
    \captionsetup{font=small} \includegraphics[width=\linewidth]
    {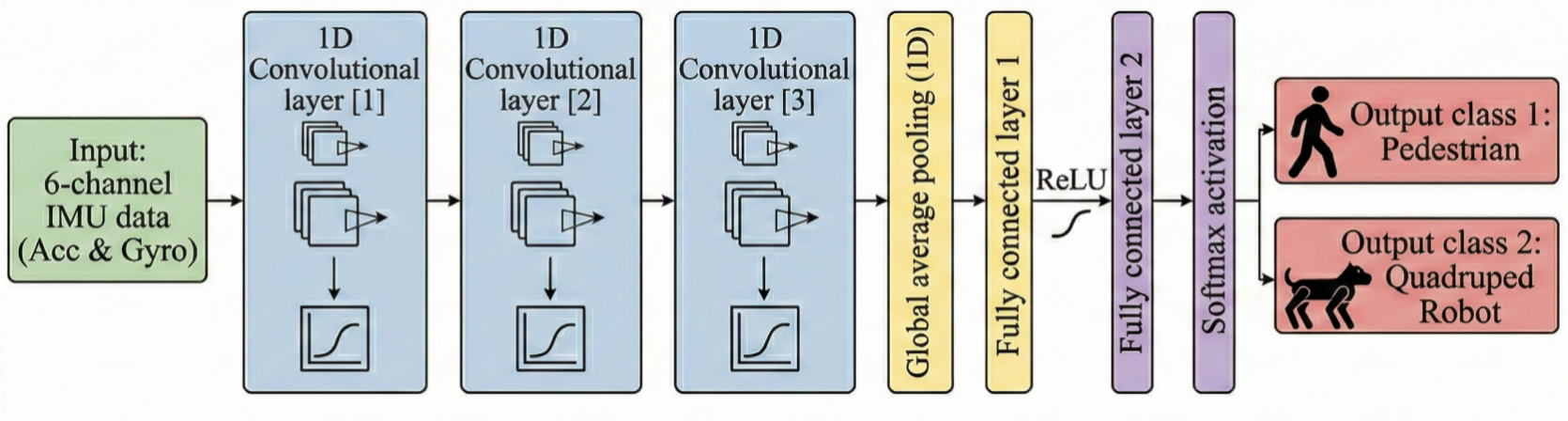}
    \caption{Architecture of the 1-D CNN classifier used in the rule-based expert selection network.}
    \label{fig:classifier}
\end{figure}

All convolutional layers use Kaiming normal initialization to ensure stable signal propagation and prevent gradient vanishing or explosion.
After the final convolutional layer, adaptive average pooling produces a fixed-length feature vector, which is then fed into a fully connected layer to generate the prediction logits for two classes. The detailed architecture of the 1-D CNN classifier is illustrated in Fig.~\ref{fig:classifier}.

\textbf{Expert selection module.}
A lightweight selection module is implemented based on the output of the classification network. The network generates prediction logits that are converted into probabilities via a softmax function, yielding a platform category and its confidence score.
A simple if–else rule then directs each IMU sequence to its corresponding pre-trained displacement estimation expert network.
The classifier is trained in advance using labeled inertial data from the platform categories to be recognized and is optimized with the standard Cross-Entropy loss\cite{mao2023cross}. Extending the framework to new platforms requires adding representative training data, retraining the selector, and learning a corresponding expert.
This design is computationally efficient and provides an effective means for generalizing inertial odometry across heterogeneous motion platforms.
\subsection{Displacement Prediction Network}
\label{sec:displacement}
Inertial odometry regresses motion states from sequential IMU measurements.
From a physical perspective, accelerometer readings contain gravity-related components and motion-induced specific force, while gyroscope readings describe the rotational dynamics of the carrier. These quantities are not independent: roll and pitch affect the projection of gravity onto each accelerometer axis, and are in turn correlated with angular velocities through the platform motion.
Since accelerations and angular velocities along each axis are correlated yet heterogeneous, capturing both temporal and inter-axis dependencies is essential for accurate motion estimation. Inspired by Crossformer~\cite{Crossformer}, we design a dual-stage attention enhanced motion feature encoder–decoder architecture to jointly learn these dependencies.

\textbf{Embedding layer.} Given an inertial sequence $\boldsymbol{\mathcal{X}} \in \mathbb{R} ^{6 \times L }$ of length $L = 200$, we first partition it into equal sub-segments of length $L_{seg} = 10$ for each dimension. The $i$-th segment of the $j$-th dimension, $\boldsymbol{\mathcal{X}}^{seg}_{i,j}$, is linearly projected and augmented with a learnable positional embedding:
\begin{equation}
    \boldsymbol{h}_{i,j} = \boldsymbol{H\mathcal{X}}^{seg}_{i,j}+\boldsymbol{E}^{pos}_{i,j}
\end{equation}
where $\boldsymbol{H}\in \mathbb{R} ^{d_{model} \times L_{seg} }$ is the projection matrix, $\boldsymbol{E}^{pos}_{i,j}\in \mathbb{R}^{d_{model}}$ is the positional encoding, and $d_{model} = 256$ denotes the embedding size.

\textbf{Motion Feature Encoder with Dual-Stage Attention.} The resulting 2D feature tensor $\boldsymbol{Y} \in \mathbb{R}^{N \times D \times d_{model}}$ (with $N = 20$ segments and $D = 6$ dimensions) is processed through two successive attention stages:
\begin{itemize}

    \item Temporal Self-Attention: To capture long-range temporal dependencies within each sensor dimension, the feature slice corresponding to the $d$-th dimension $Y_{:,d} \in \mathbb{R}^{N \times d_{model}}$ is processed by a temporal self-attention mechanism:
    \begin{equation}
        Attention(Q,K,V)=softmax\left ( \frac{QK^{T}}{\sqrt{d_{k}} }  \right ) V
    \end{equation}
    where
    \begin{equation}
        Q=Y_{:,d}W_{Q}, K=Y_{:,d}W_{K}, V=Y_{:,d}W_{V}
    \end{equation}
    $W_{Q}, W_{K}, W_{V} \in \mathbb{R}^{d_{k} \times d_{model}}$ are learnable projection matrices. $d_k=\frac{d_{model}}{h}$ denotes the dimension of each attention head, with $h$ being the number of heads. This operation models temporal correlations within each dimension independently.
    \item Dimensional Self-Attention: After temporal encoding, multi-head self-attention is further performed across dimensions to model inter-dimensional dependencies among sensor channels. For each temporal position $n$, the corresponding feature vector $Y_{n,:} \in \mathbb{R}^{D \times d_{model}}$ is processed by a dimensional self-attention mechanism. A lightweight router~\cite{Crossformer} is introduced to selectively aggregate the most relevant inter-dimensional dependencies, reducing computational overhead while preserving representational richness.
\end{itemize}

This dual-stage attention mechanism yields a representation $\boldsymbol{Z} \in \mathbb{R}^{N \times D \times d_{model}}$ that effectively encodes both temporal and cross-dimensional motion dynamics for inertial odometry estimation.

The encoder consists of $N$ layers, each merging every two adjacent temporal features (except for the first layer) to obtain coarser temporal representations. Each layer applies the dual-stage attention module to extract multi-scale temporal and dimensional features, enabling hierarchical motion understanding.
After encoding, we obtain multi-resolution representations $\boldsymbol{Y}^{enc, 0}$, $\boldsymbol{Y}^{enc, 1}$, ..., $\boldsymbol{Y}^{enc, N}$, where higher layers capture long-term dynamics and lower layers retain fine-grained motion details.

\textbf{Multiscale Feature Decoder with Cross-Attention.} The decoder mirrors the encoder with $N+1$ layers. The input to the initial decoder layer is a learnable positional embedding $E^{dec}$, while subsequent layers take the output of the preceding decoder layer as input. Each layer first processes its input through the dual-stage attention module and then performs cross-attention with the corresponding encoder features. The fused representation is refined via an MLP and layer normalization, and the final decoder output is passed through a fully connected layer to regress the 3D displacement and its associated covariance.

\textbf{Uncertainty-aware training with Huber–Gaussian loss.} To achieve robust regression with uncertainty modeling, we employ a Huber-Gaussian loss proposed in \cite{AirIO}:
\begin{equation}
    \mathcal{L} = \mathcal{L}_{Huber} + \lambda\mathcal{L}_C
\end{equation}
where the $\lambda = 1e^{-4}$. The Huber function is defined as
\begin{equation}
    \mathcal{L}_{Huber} = \begin{cases}
\frac{1}{2}(\boldsymbol{\hat{d}}_{i} - \boldsymbol{d}_{i})^2 & \text{if } |\boldsymbol{\hat{d}}_{i} - \boldsymbol{d}_{i}| \leq \delta \\
\delta \cdot|\boldsymbol{\hat{d}}_{i} - \boldsymbol{d}_{i}| - \frac{1}{2}\delta^2 & \text{otherwise}
\end{cases}
\end{equation}
\begin{equation}
\mathcal{L}_{C} = \frac{1}{2} (\boldsymbol{\hat{d}}_{i} - \boldsymbol{d}_{i})\hat{\Sigma} _{i}^{-1}(\boldsymbol{\hat{d}}_{i} - \boldsymbol{d}_{i})^T + \frac{1}{2}ln(det\hat{\Sigma}_i) 
\end{equation}
where $\delta = 0.005$, $\hat{d}_i$ and $d_i$ denote the predicted and ground-truth displacements, respectively, and $\hat{\Sigma}_i \in \mathbb{R}^{3\times3}$ denotes the covariance matrix predicted by the network for the $i$-th sample.

This formulation provides robustness to outliers via the Huber loss, while the Gaussian likelihood term enables uncertainty-aware weighting of predictions. The Huber-Gaussian loss approximates the Mean Squared Error (MSE) for small errors, ensuring fitting accuracy, and transitions to a linear penalty for large errors, effectively mitigating the impact of outliers. This characteristic contributes to more stable training and better generalization in cross-platform scenarios. Compared to the loss function used in \cite{TLIO}, the Huber-Gaussian loss simultaneously optimizes both the size of prediction errors and the accuracy of uncertainty estimation, leading to more stable statistical consistency of the uncertainty (Section~\ref{sec:Uncertainty}).

\subsection{EKF-Based State Estimation}
We employ the Extended Kalman Filter (EKF) for state estimation, integrating the output of the displacement prediction network with the state optimization process.
\label{sec:EKF}
\subsubsection{State Model}
The filter state vector is defined as: $\boldsymbol{S} = (\boldsymbol{\lambda_1}, \dots, \boldsymbol{\lambda_n}, \boldsymbol{x})$. The historical state vector $\boldsymbol{\lambda}_{i}(i=1, \ldots, n)$ contains pose information from $n$ past moments, and $\boldsymbol{x}=\left(\mathrm{_{i}^{w}R},\mathrm{^{w}}\boldsymbol{v},\mathrm{^{w}}\boldsymbol{p}, b_{g}, b_{a}\right)$ represents the current state. Where $\mathrm{_{i}^{w}}\mathrm{R} \in SO(3)$ is the rotation matrix from the IMU frame to the world frame; $\boldsymbol{\mathrm{^{w}}v}$ and $\boldsymbol{\mathrm{^{w}}p}$ are the velocity and position in the world frame; $b_{g} $ and $b_{a} $ are the biases of the gyroscope and accelerometer. The EKF uses an error state representation:
\begin{equation}
    \tilde{\boldsymbol{\lambda}}_{i}=(\tilde{\boldsymbol{\theta}}_{i}, \delta {\tilde{\boldsymbol{p}}_{i}})
\end{equation}
\begin{equation}
    \tilde{\boldsymbol{x}}=(\tilde{\boldsymbol{\theta}}, \delta{\tilde{\boldsymbol{v}}}, \delta{\tilde{\boldsymbol{p}}}, \delta{\tilde{b}_{g}}, \delta{\tilde{b}_{a}})
\end{equation}
where $\tilde{\boldsymbol{\theta}}=\log _{SO3}(\mathrm{R} \hat{\mathrm{R}}^{-1})$ represents the rotational error. $\delta{\tilde{\boldsymbol{v}}}, \delta{\tilde{\boldsymbol{p}}}, \delta{\tilde{b}_{g}}, \delta{\tilde{b}_{a}}$ represent the errors in velocity, position, and bias, respectively.

\subsubsection{State Propagation}
Extended Kalman Filter propagates the current state $\boldsymbol{x}$ using inertial data and a kinematic model:
\begin{equation}
\mathrm{^{w}_{i}\hat{R}_{n+1}}=\mathrm{_{i}^{w}\hat{R}_{n}}\exp _{S O 3}((\boldsymbol{\omega}_{n}-\hat{\boldsymbol{b}}_{g_{n}})\Delta t)
\end{equation}
\begin{equation}
^{\mathbf{w}}\hat{\boldsymbol{v}}_{n+1}=\mathbf{^{w}}\hat{\boldsymbol{v}}_{n}+\mathbf{^{w}}\boldsymbol{g} \Delta t+\mathbf{_{i}^{w}} \hat{\mathbf{R}}_{n}(\boldsymbol{a}_{n}-\hat{\boldsymbol{b}}_{a_{n}})\Delta t
\end{equation}
\begin{equation}
^{\mathbf{w}} \hat{\boldsymbol{p}}_{n+1}=\mathbf{^{w}} \hat{\boldsymbol{p}}_{n}+\frac{1}{2}\Delta t(^{\mathbf{w}}\hat{\boldsymbol{v}}_{n+1} + \mathbf{^{w}}\hat{\boldsymbol{v}}_{n})
\end{equation}
\begin{equation}
\hat{b}_{g_{(n+1)}}=\hat{b}_{g_{n}}+n_{g_{n}}
\end{equation}
\begin{equation}
\hat{b}_{a_{(n+1)}}=\hat{b}_{a_{n}}+n_{a_{n}}
\end{equation}
where $\exp _{SO3}$ is the exponential map for the rotation group, and $\Delta t$ is the sampling period of the inertial data. $\boldsymbol{g}$ represents the gravity acceleration in the world frame. $n$ donates Gaussian-distributed noise.

\subsubsection{State Update}
The measurement function $h(X)$ is:
\begin{equation}
h(\boldsymbol{X}) =\mathbf{R}^T(\mathbf{^{w}}\boldsymbol{p}_j-\mathbf{^{w}}\boldsymbol{p}_i)=\boldsymbol{\hat{d}}_{ij} + n_{ij}
\end{equation}
where $\mathbf{R}^T$ represents the transformation that converts the displacement difference in the world frame to the local frame. Using the measurement Jacobian matrix $\mathbf{H}$ and the covariance $\hat{\boldsymbol{\Sigma}}_{ij}$ output by the neural network, the Kalman gain $\mathbf{K}$ is calculated as:
\begin{equation}
\mathbf{K} = \boldsymbol{P}\mathbf{H}^T(\mathbf{H}\boldsymbol{P}\mathbf{H}^T + \hat{\boldsymbol{\Sigma}}_{ij})^{-1}
\end{equation}
The state $\mathbf{S}$ is then updated by
\begin{equation}
\mathbf{S} \leftarrow \mathbf{S} \oplus \left(\mathbf{K}(h(\mathbf{S}) - \hat{\boldsymbol{d}}_{ij})\right)
\end{equation}
where $\oplus$ denotes the addition operation on the $SO(3)$ group when updating pose state variables, while other state variables use ordinary addition. The covariance is updated following the standard EKF correction step.

\section{EXPERIMENT}

\begin{figure*}[!t]
    \vspace{-3pt}
    \captionsetup{font=small,skip=3pt}
    \setlength{\fboxsep}{2pt}
    \includegraphics[width=\linewidth]{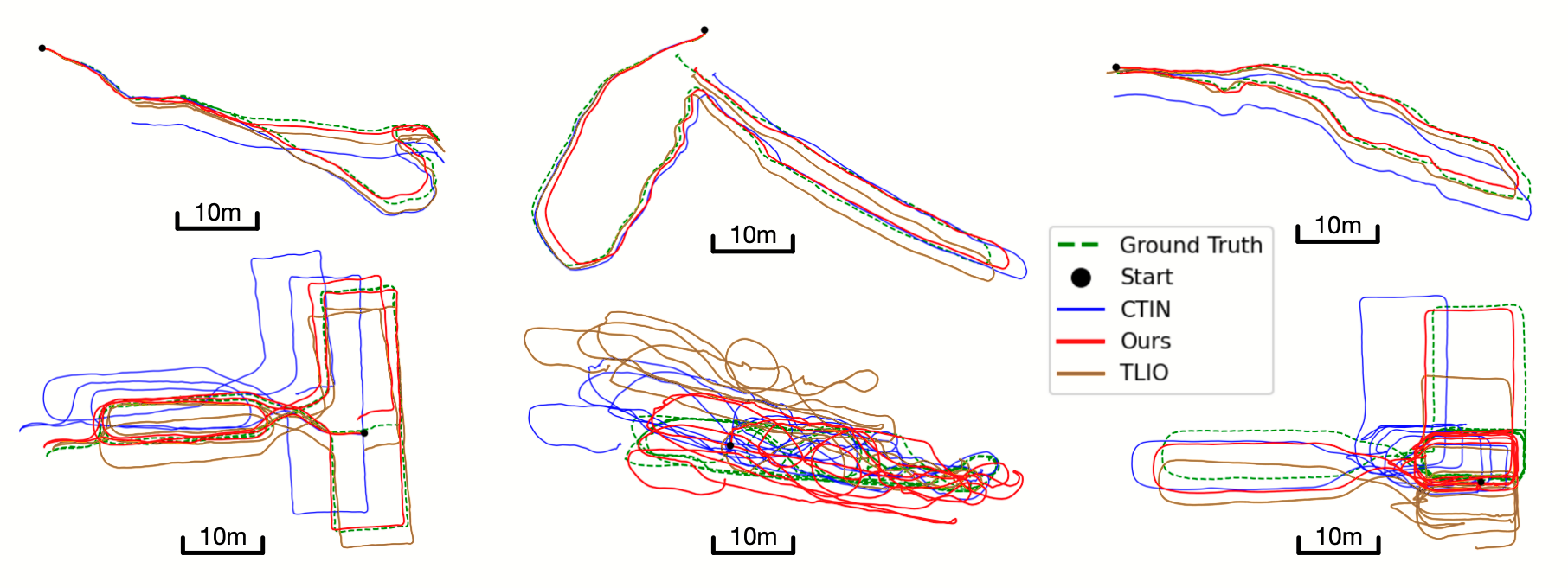}
    \caption{Trajectory comparisons of representative experimental results. The top row corresponds to GrandTour test sequences, and the bottom row corresponds to RoNIN test sequences.}
    \label{fig:trajectory}
    \vspace{-4pt}
\end{figure*}

\subsection{Experiment Setup}
\subsubsection{Datasets}
We evaluated the proposed method on both pedestrian and quadruped robot inertial datasets.

For pedestrians, we
use the RoNIN\cite{RoNIN} dataset, a large-scale inertial odometry benchmark collected with handheld smartphones.
The dataset contains 42.7 hours of IMU measurements with corresponding ground-truth trajectories. We follow the official training, validation, and test splits.
The test set is further divided into \emph{Seen} and \emph{Unseen} subsets, each comprising 32 trajectories. The \emph{Seen} subset includes trajectories from subjects present in the training set, whereas the \emph{Unseen} subset contains trajectories from previously unseen subjects.

For quadruped robots, we use the public GrandTour dataset\cite{grandtour}, which provides synchronized multimodal measurements collected from quadruped platforms in real-world environments. In this work, we use the IMU data from the NovAtel CPT7 device as the inertial input and the corresponding odometry trajectories as ground truth. 
After data screening, 36 trajectories with complete and reliable IMU and ground-truth data are retained, while the remaining sequences are excluded due to missing ground-truth trajectories. The selected subset is split into 24 training, 6 validation, and 6 test sequences.

In addition to the public dataset, we collected a quadruped inertial dataset using a Unitree Go2 robot. As shown in Fig.~\ref{fig:frame}, an iPhone 14 Pro Max was rigidly mounted on the robot to record IMU data, while the reference trajectory was obtained from onboard LiDAR odometry. The phone’s IMU provided measurements at 100 Hz, which were resampled to 200 Hz along with the ground-truth trajectories using linear interpolation. The dataset consists of 30 sequences with a total duration of approximately 1.5 hours and is used for additional validation under a different platform and sensor configuration. Each sequence lasts approximately 3 minutes, with traveled distances ranging from 50 to 100~m. During steady forward locomotion, the nominal speed is approximately 0.46~m/s.

\begin{table*}[!ht]
  \centering
  \captionsetup{font=small}
  \caption{Results comparison of different models on RoNIN, GrandTour and Go2 dataset.}
  \label{tab:results}
  \small
  \renewcommand{\arraystretch}{1.2}
  \setlength{\tabcolsep}{9.5pt}
  \begin{tabular}{c|cc|cc|cc|cc}
    \toprule
    \multirow{2}{*}{Model} 
    & \multicolumn{2}{c|}{RoNIN\_seen} 
    & \multicolumn{2}{c|}{RoNIN\_unseen} 
    & \multicolumn{2}{c|}{GrandTour} 
    & \multicolumn{2}{c}{Go2} \\
    \cmidrule(lr){2-3} \cmidrule(lr){4-5} \cmidrule(lr){6-7} \cmidrule(lr){8-9}
    & ATE (m) 
    & RTE (m) 
    & ATE (m) 
    & RTE (m) 
    & ATE (m) 
    & RTE (m) 
    & ATE (m) 
    & RTE (m) \\
    \midrule
    RoNIN-ResNet & \cellcolor{green!35}4.11 & \cellcolor{green!30}3.28 & \cellcolor{green!35}5.26 & \cellcolor{green!35}4.62 & \cellcolor{green!35}6.51 & \cellcolor{green!35}5.32 & \cellcolor{green!05}3.62 & \cellcolor{green!20}1.95 \\
    RoNIN-TCN    & \cellcolor{green!30}4.31 & \cellcolor{green!10}5.02 & \cellcolor{green!15}6.74 & \cellcolor{green!15}5.10 & \cellcolor{green!05}8.21 & \cellcolor{green!00}7.39 & \cellcolor{green!10}3,61 & \cellcolor{green!00}2.46 \\
    RONIN-LSTM   & \cellcolor{green!10}6.19 & \cellcolor{green!15}4.56 & \cellcolor{green!10}7.61 & \cellcolor{green!10}5.87 & \cellcolor{green!00}8.56 & \cellcolor{green!10}7.23 & \cellcolor{green!00}3.98 & \cellcolor{green!05}2.08 \\
    TLIO         & \cellcolor{green!15}4.74 & \cellcolor{green!20}3.96 & \cellcolor{green!25}6.11 & \cellcolor{green!30}4.75 & \cellcolor{green!40}6.09 & \cellcolor{green!40}4.63 & \cellcolor{green!35}2.22 & \cellcolor{green!40}1.37 \\
    IMUNet       & \cellcolor{green!05}6.56 & \cellcolor{green!00}5.90 & \cellcolor{green!05}8.35 & \cellcolor{green!05}6.02 & \cellcolor{green!10}8.03 & \cellcolor{green!05}7.32 & \cellcolor{green!30}3.17 & \cellcolor{green!30}1.77 \\
    MobileNet    & \cellcolor{green!00}6.89 & \cellcolor{green!05}5.09 & \cellcolor{green!00}8.80 & \cellcolor{green!00}6.96 & \cellcolor{green!15}7.53 & \cellcolor{green!15}6.48 & \cellcolor{green!15}3.56 & \cellcolor{green!10}2.07 \\
    MnasNet      & \cellcolor{green!40}3.96 & \cellcolor{green!40}2.80 & \cellcolor{green!30}6.08 & \cellcolor{green!20}5.03 & \cellcolor{green!25}6.77 & \cellcolor{green!20}6.04 & \cellcolor{green!25}3.45 & \cellcolor{green!25}1.93 \\
    EfficientNet & \cellcolor{green!20}4.59 & \cellcolor{green!25}3.46 & \cellcolor{green!20}6.25 & \cellcolor{green!25}4.84 & \cellcolor{green!20}6.89 & \cellcolor{green!25}5.68 & \cellcolor{green!40}2.18 & \cellcolor{green!35}1.43 \\
    CTIN         & \cellcolor{green!25}4.34 & \cellcolor{green!35}3.08 & \cellcolor{green!40}5.25 & \cellcolor{green!40}4.54 & \cellcolor{green!30}6.67 & \cellcolor{green!30}5.65 & \cellcolor{green!20}3.54 & \cellcolor{green!15}2.03 \\
    \textbf{Ours} & \cellcolor{green!50}\textbf{3.10} & \cellcolor{green!50}\textbf{2.42} & \cellcolor{green!50}\textbf{4.79} & \cellcolor{green!50}\textbf{4.05} & \cellcolor{green!50}\textbf{5.37} & \cellcolor{green!50}\textbf{4.18} & \cellcolor{green!50}\textbf{1.03} & \cellcolor{green!50}\textbf{0.84} \\
    \bottomrule
  \end{tabular}
\end{table*}

\subsubsection{Baseline}
We compared our method with a wide range of state-of-the-art inertial odometry approaches, including:

\begin{itemize}
    \item End-to-end Inertial Odometry: RoNIN\_TCN\cite{RoNIN}, RoNIN\_LSTM\cite{RoNIN}, RoNIN\_ResNet\cite{RoNIN}.
    \item Transformer-based Inertial Odometry: CTIN\cite{CTIN}.
    \item EKF-integrated Inertial Odometry: TLIO\cite{TLIO}.
    \item Lightweight Inertial Odometry: IMUNet\cite{IMUNet}, EfficientNet\cite{IMUNet}, MobileNet\cite{IMUNet}, MnasNet\cite{IMUNet}.
\end{itemize}

\subsubsection{Metrics}
We evaluated performance using Absolute Trajectory Error (ATE) and Relative Trajectory Error (RTE).
ATE measures global trajectory accuracy as the root-mean-square distance between aligned estimated and ground-truth positions.
RTE quantifies local drift by comparing estimated and ground-truth displacements within 1-minute time windows.

\subsubsection{Training and Inference}
For a fair comparison, all learning-based baseline methods were retrained on the training split of the corresponding target dataset before evaluation.
We adopted a sliding-window strategy, segmenting 200 Hz IMU data into 1-second windows. Training for all models used the Adam optimizer with a batch size of 128, a learning rate of $1\text{e-}4$, and up to 100 epochs. All experiments were conducted on an NVIDIA RTX 4090 GPU.
Compared with one end-to-end generalist model, X-IONet activates only a single expert network for each input window during inference, resulting in efficient deployment.

\subsection{Evaluation on Pedestrian Datasets}

Table~\ref{tab:results} summarizes the performance of all compared methods on the RoNIN pedestrian dataset. X-IONet achieves state-of-the-art performance under both Seen and Unseen testing conditions. Specifically, compared with the strongest baseline, X-IONet reduces ATE and RTE by 14.3\% and 11.4\%, respectively. These results confirm that the cross-platform design of X-IONet does not compromise accuracy on pedestrian scenarios, effectively capturing human gait patterns without overfitting to specific motion styles.

Fig.~\ref{fig:trajectory} (left panel) presents representative trajectories for the RoNIN test sequences. Only two representative baseline methods are visualized: CTIN\cite{CTIN}, which employs a Transformer-based architecture, and TLIO\cite{TLIO}, which uses an EKF framework. The trajectories estimated by X-IONet remain consistently closer to the ground truth compared to the baselines, qualitatively supporting the quantitative improvements reported in Table~\ref{tab:results}.

\subsection{Evaluation on Quadruped Robot Datasets}

\begin{figure}[!t]
    \centering
    \captionsetup{font=small}    \includegraphics[width=\linewidth]{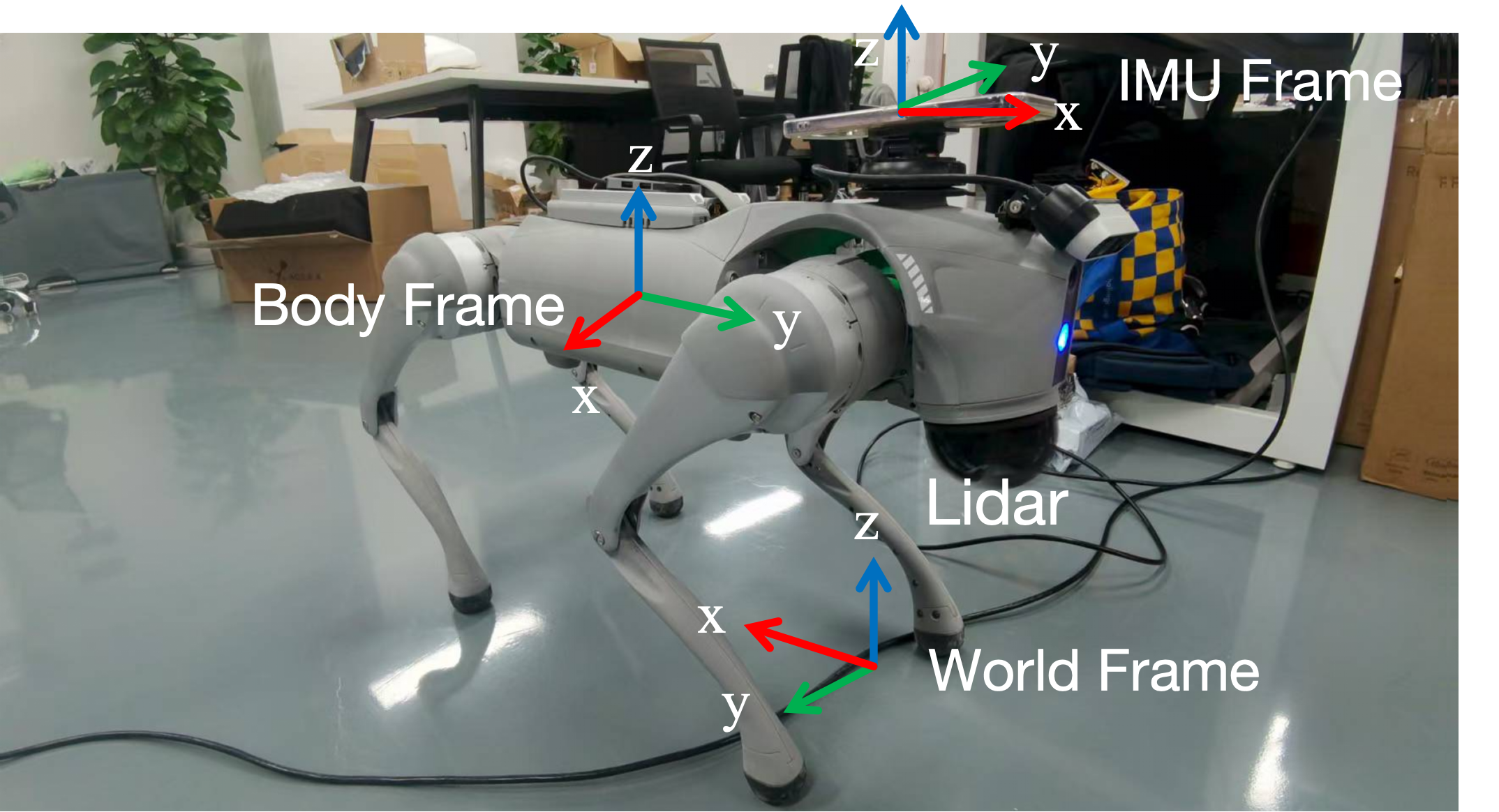}
    \caption{The quadruped robot used in the experiments.}
    \label{fig:frame}
\end{figure}

\begin{figure}[!t]
    \centering    \captionsetup{font=small}
    \includegraphics[width=\linewidth]{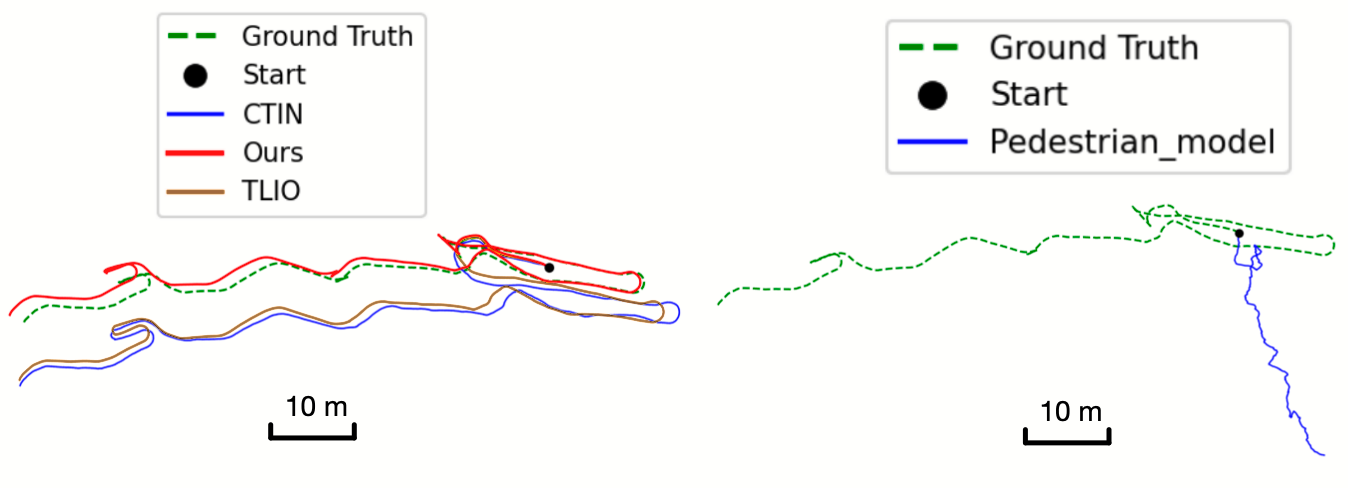}
    \caption{Trajectory comparison on a representative self-collected Go2 test sequence (left), and the trajectory of the quadruped robot predicted using the pedestrian model (right).}
    \label{fig:Go2_trajectory}
\end{figure}

Quadruped robot evaluation is performed on two datasets: the public GrandTour dataset and our self-collected Go2 dataset. Table~\ref{tab:results} reports the corresponding results. On the GrandTour dataset, X-IONet achieves the best overall performance, reducing ATE and RTE by 11.8\% and 9.7\% compared with the strongest baseline. These results indicate that X-IONet can accurately model quadruped inertial dynamics rather than relying on patterns learned from pedestrian motion.
For our self-collected Go2 dataset, the performance gains are even more pronounced. X-IONet reduces ATE by 52.8\% and RTE by 41.3\% relative to the strongest baseline. These improvements demonstrate that the proposed framework is robust to different quadruped platforms and sensor configurations.
Fig.~\ref{fig:frame} shows the quadruped robot used in the experiments, while Fig.~\ref{fig:Go2_trajectory} presents a representative trajectory comparison on a Go2 test sequence. As in the GrandTour evaluation, CTIN and TLIO are included as reference baselines. The trajectories estimated by X-IONet remain closest to the ground truth, visually confirming the quantitative advantages reported in Table~\ref{tab:results}.

Overall, these evaluations demonstrate that X-IONet consistently generalizes across pedestrian and quadruped scenarios, including both public benchmarks and a distinct real-world robot platform. The method effectively captures platform-specific motion dynamics while maintaining robustness across different sensor setups and locomotion modalities.

\subsection{Ablation Study}
We conducted ablation experiments to quantify the contribution of each key component in X-IONet. The following variants were evaluated:

\begin{itemize}
    \item \textbf{W/O Dual-Stage Attention Mechanism:} Removing dimensional attention and retaining only temporal attention significantly degrades performance.
    \item \textbf{W/O Hierarchical encoder-decoder architecture:} Eliminating the hierarchical structure forces the decoder to attend only to the final encoder layer.
    \item \textbf{W/O Extended Kalman Filter:} Using raw network outputs without EKF refinement leads to larger drift.
    \item \textbf{W/O Uncertainty in Loss Function:} Replacing the Huber–Gaussian loss with a standard Mean Squared Error (MSE) ignores uncertainty learning.
\end{itemize}

The results in Table~\ref{tab:Ablation_study} show the quantitative impact of each component across the RoNIN, GrandTour, and Go2 datasets. Removing the Dual-Stage Attention Mechanism consistently degrades performance, with ATE and RTE increasing by 33.4\%–107.8\% and 34.6\%–85.7\%, respectively. This highlights the importance of jointly modeling temporal dependencies and cross-dimensional correlations for accurate inertial odometry across pedestrian and quadruped scenarios.

Ablating the hierarchical encoder-decoder architecture leads to ATE and RTE increases of 7.6\%–40.8\% and 8.6\%–20.2\%, demonstrating that multi-scale feature aggregation improves the network’s ability to capture both fine-grained motion details and long-range temporal context.

Removing the EKF refinement results in ATE and RTE increases of 12.1\%–32.0\% and 11.2\%–15.7\%, confirming that EKF-based integration reduces drift and enhances trajectory accuracy over time.
Finally, removing uncertainty-awareness from the loss function increases ATE and RTE by 16.7\%–69.9\% and 24.4\%–61.9\%, respectively, indicating that modeling prediction uncertainty significantly improves robustness across diverse motion patterns and sensing conditions.

\begin{table}[!t]
  \centering
  \small
  \renewcommand{\arraystretch}{1.3}
  \setlength{\tabcolsep}{1.5pt}
  \captionsetup{font=small}
  \caption{Ablation study on RoNIN, GrandTour and Go2 datasets.}
  \label{tab:Ablation_study}
  \newcolumntype{L}{>{\raggedright\arraybackslash}X}
  \newcolumntype{C}{>{\centering\arraybackslash}p{20pt}}
  \begin{tabularx}{\linewidth}{L | C C | C C | C C}
    \toprule
    \textbf{Model Variation} & \multicolumn{2}{c|}{\textbf{RoNIN}} & \multicolumn{2}{c|}{\textbf{GrandTour}} & \multicolumn{2}{c}{\textbf{Go2}} \\
    \cmidrule(lr){2-3} \cmidrule(lr){4-5} \cmidrule(lr){6-7}
    & ATE (m)& RTE (m)& ATE (m) & RTE (m) & ATE (m) & RTE (m) \\
    \midrule
    \textbf{Ours} & \textbf{3.95} & \textbf{3.24} & \textbf{5.37} & \textbf{4.18} & \textbf{1.03} & \textbf{0.84} \\
    \midrule
    
    \textit{w/o} Dual-Stage Attention   
        & 5.27
        & 4.36
        & 7.61 
        & 6.32
        & 2.14 
        & 1.56 \\
        
    \textit{w/o} Hierarchical Encoder-Decoder  
        & 4.37
        & 3.52
        & 5.78 
        & 4.87
        & 1.45
        & 1.01\\
        
    \textit{w/o} Extended Kalman Filter 
        & 4.46
        & 3.75
        & 6.02
        & 4.65 
        & 1.36 
        & 0.97 \\
        
    \textit{w/o} Loss Uncertainty
        & 4.61
        & 4.03
        & 7.19 
        & 5.87 
        & 1.75
        & 1.36 \\
    \bottomrule
  \end{tabularx}
\end{table}

\subsection{Why the Huber–Gaussian loss?}
Compared to the staged "MSE + maximum-likelihood" training used in TLIO~\cite{TLIO}, the Huber–Gaussian loss enhances robustness, stability, and uncertainty accuracy. The Huber term suppresses outlier influence and prevents the model from artificially inflating uncertainty, while the Gaussian likelihood maintains statistical consistency between predicted uncertainty and true error.

\label{sec:Uncertainty}
\begin{table}[!t]
    \centering
    \captionsetup{font=small}
    \caption{Comparison of the consistency of learned covariance: the 3$\sigma$ coverage is the proportion of samples where the actual error lies within three times the standard deviation of the estimated uncertainty.}
    \label{tab:Ablation 3}
    \footnotesize
    \renewcommand{\arraystretch}{1.2}
    \setlength{\tabcolsep}{10pt}
    \begin{tabular}{lc}
        \toprule
        \textbf{Loss Function} & \textbf{3$\sigma$ Coverage (\%)} \\
        \midrule
        Huber--Gaussian Loss & 99.2\% \\
        Loss Function in TLIO\cite{TLIO} & 98.7\% \\
        \bottomrule
    \end{tabular}
\end{table}

Table~\ref{tab:Ablation 3} compares uncertainty consistency under both loss functions. The TLIO loss is more sensitive to outliers (e.g., sudden IMU disturbances), causing more samples to exceed the $3\sigma$ bound. In contrast, the Huber term attenuates outlier effects, improving coverage by keeping more samples within the $3\sigma$ range. This demonstrates that the Huber–Gaussian loss yields more stable and reliable uncertainty estimates.

\begin{figure}[h]
    \centering    \captionsetup{font=small}\includegraphics[width=0.88\linewidth]{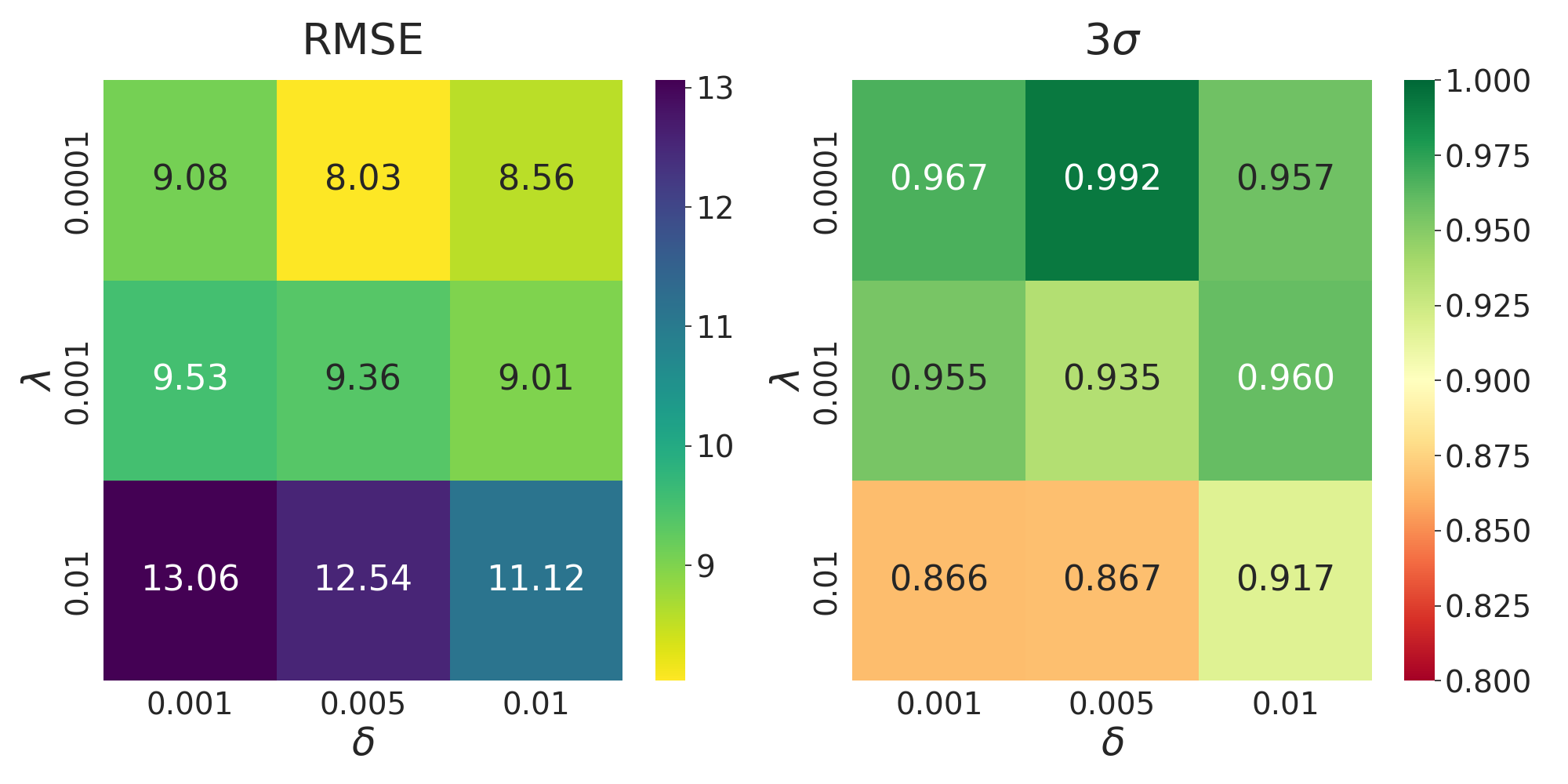}
    \caption{Parameter sensitivity analysis of the Huber-Gaussian loss on the GrandTour dataset.}
    \label{fig:heatmap}
\end{figure}

To further justify the choice of the Huber--Gaussian loss parameters, we evaluated a grid of Huber thresholds and Gaussian regularization weights, with $\delta \in \{0.001, 0.005, 0.01\}$ and $\lambda \in \{0.0001, 0.001, 0.01\}$. As shown in Fig.~\ref{fig:heatmap}, we measured the RMSE of the predicted displacement and the $3\sigma$ coverage. Among all tested settings, $\delta=0.005$ and $\lambda=0.0001$ achieved the best overall performance.

\subsection{Effectiveness of the Rule-Based Expert Selection Network}

\begin{table}[h]
    \centering
    \captionsetup{font=small}
    \setlength{\tabcolsep}{5pt}
    \renewcommand{\arraystretch}{1.5}
    \caption{Classification accuracy of the rule-based expert selection network evaluated on 1-second windows from the RoNIN pedestrian dataset and the GrandTour quadruped robot dataset.}
    \label{tab:classification_accuracy} 
    \begin{tabular}{lccc}
        \toprule
        \textbf{Dataset} & \textbf{Target Class} & \textbf{Number of Samples} & \textbf{Accuracy} \\
        \midrule
        RoNIN & Pedestrian & 376,262 & 99.72\% \\
        GrandTour & Quadruped Robot & 40,445 & 100.00\% \\
        \bottomrule
    \end{tabular}
    \arrayrulecolor{black}
    \color{black}
\end{table}

Human and quadruped motion patterns differ substantially in dynamics and inertial characteristics, causing models trained on one domain to fail on the other. As shown in Fig.~\ref{fig:Go2_trajectory}, applying a human-trained model to quadruped data results in severe trajectory distortion.
We tested the classifier on 376,262 one-second windows from the pedestrian RoNIN dataset and 40,445 windows from the quadruped robot GrandTour dataset. As summarized in Table~\ref{tab:classification_accuracy}, the classifier achieved accuracies of 99.72\% and 100.00\%, respectively, demonstrating highly reliable platform recognition and expert routing performance.

\begin{figure}[h]
    \centering   \captionsetup{font=small}\includegraphics[width=\linewidth]{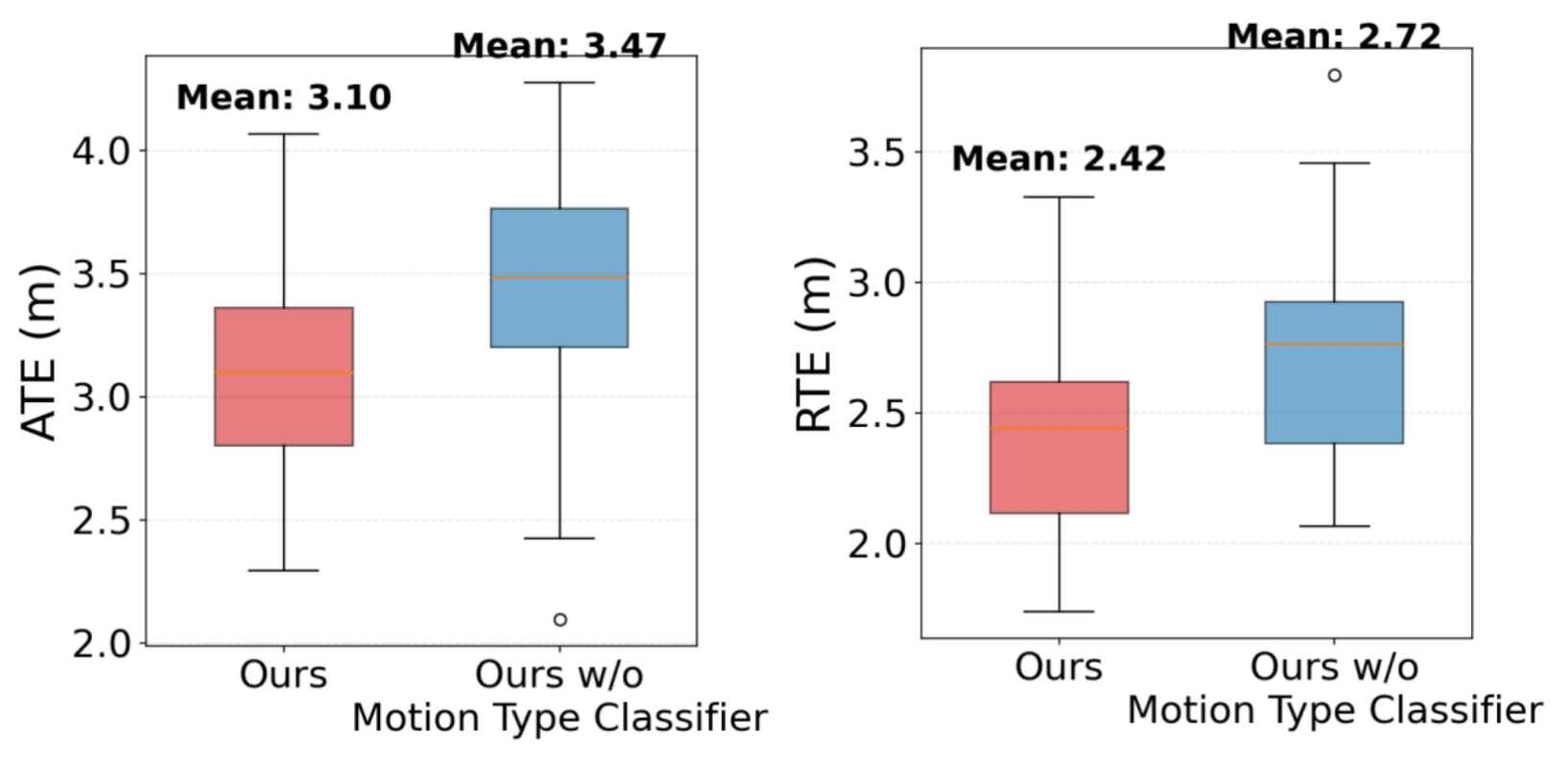}
    \caption{Boxplot comparison of ATE and RTE for X-IONet with and without the motion type classifier. The mean values are annotated above each boxplot.}
    \label{fig:boxplots}
\end{figure}

We evaluate whether direct mixed-domain training can replace expert selection. After jointly training on pedestrian and quadruped data, performance on pedestrian tests still drops—ATE and RTE degrade by 11.9\% and 12.4\%, respectively (Fig.~\ref{fig:boxplots}). These results indicate that naive joint training cannot achieve true cross-platform generalization. In contrast, the proposed rule-based classifier effectively routes IMU data to specialized experts, providing a simple and reliable mechanism for cross-platform inertial odometry.

\section{CONCLUSIONS}
This paper presented X-IONet, a cross-platform inertial odometry framework. 
The displacement prediction network employs a dual-stage attention architecture to jointly capture long-range temporal dependencies and inter-axis correlations from raw inertial signals.
The model regresses displacement and uncertainty, which are refined with an EKF for robust state estimation.
Extensive experiments on both the RONIN pedestrian dataset and a newly collected Unitree Go2 quadruped dataset demonstrated state-of-the-art performance. X-IONet reduced ATE and RTE by 14.3\% and 11.4\% on RoNIN, and by 52.8\% and 41.3\% on Go2, compared with the best existing methods. Ablation studies confirmed the effectiveness of the dual-stage attention mechanism, 
the hierarchical encoder-decoder architecture, and the EKF refinement.
In addition, a rule-based front-end classification module enabled robust cross-platform generalization across humans and robots.

The current evaluation is limited to trained pedestrian and quadruped domains; performance on other platforms, unseen robots, and heterogeneous IMU configurations remains to be explored. Future work will extend the framework to a broader range of platforms and improve robustness to unseen carrier instances, different robot models, and diverse IMU hardware. An important direction is to investigate whether cross-robot generalization is better achieved through jointly trained experts or through robot-specific experts combined with a more fine-grained classifier.

\bibliographystyle{IEEEtran}
\bibliography{Mybib}

\end{document}